# STRIDE: Strategy-to-Tactics Retrieval-Guided Skill Disclosure and Evolution for Agentic Reinforcement Learning

Shixin Hong
hongshixin.hsx@taobao.com
Taobao & Tmall Group of Alibaba
Hangzhou, China

Wanqing Cui
cuiwanqing.cwq@taobao.com
Taobao & Tmall Group of Alibaba
Hangzhou, China

Yibo Hu
boxuan.hyb@taobao.com
Taobao & Tmall Group of Alibaba
Hangzhou, China

Naen Xu
xunaen@zju.edu.cn
Zhejiang University
Hangzhou, China
Taobao & Tmall Group of Alibaba
Hangzhou, China

Zhiwei Huang
xuanan.hzw@alibaba-inc.com
Taobao & Tmall Group of Alibaba
Hangzhou, China

Meiguang Jin
meiguang.jmg@taobao.com
Taobao & Tmall Group of Alibaba
Hangzhou, China

Junfeng Ma
jack.majf@taobao.com
Taobao & Tmall Group of Alibaba
Hangzhou, China

## Abstract

Skill-augmented reinforcement learning enables LLM agents to reuse task-solving knowledge across episodes, but existing systems typically expose agents to a flat skill pool and update skills only from coarse, episode-level outcomes. This conflates two distinct uses of experience: a stable strategy that should frame the whole episode, and tactical know-how that should be retrieved only when the agent reaches a local execution gap. We propose **STRIDE** (**S**trategy-to-**T**actics **R**etrieval-guided sk**I**ll **D**isclosure and **E**volution), a framework that brings progressive skill disclosure into agentic reinforcement learning. STRIDE keeps the selected strategy visible as episode context, exposes a compact tactic menu, and loads full tactic content only after an explicit retrieval action. This interface turns each retrieval into an event-level credit unit: the same event provides process feedback for learning when to retrieve and supplies evidence for skill-bank evolution. On ALFWorld, STRIDE achieves a 92.7% average success rate with Qwen3-4B and a 94.9% average success rate with Qwen3-8B. On WebShop, it improves over the strongest reproduced skill-augmented baseline by 4.5 score points and 1.9 success-rate percentage points. These results show that skill reuse is most effective when persistent strategies preserve global task structure and tactics are disclosed only at local decision bottlenecks.

## 1 Introduction

Large language model (LLM) agents can now act across domains [11, 24, 39, 40]. Their ability to learn from their own execution remains more fragile: once an episode ends in success or failure, the task-solving experience behind that trajectory is easily lost [47]. Memory-based agents preserve such experience as reflections, distilled lessons, or persistent memories [8, 23, 51]. Skill-augmented reinforcement learning goes further by converting trajectories into reusable skills and training the policy to condition on a skill bank [28, 30, 36]. Recent systems also structure the skill bank so that it can be revised and optimized alongside the policy [10, 12, 22, 50].

Once experience is stored as a structured bank, the central design question shifts from storage alone to disclosure: which part of the bank should be visible at each decision? This question is especially natural for web-facing agents, where shopping and information-discovery tasks require reusable experience across heterogeneous pages but each decision usually depends on a small, state-specific piece of guidance [39, 40]. In these tasks, an agent repeatedly moves between the user's global intent, such as satisfying product constraints or gathering evidence for a question, and local page cues such as filters, attributes, snippets, or navigation options. Injecting the entire skill bank at every page can dilute the current observation, while retrieving after every click turns memory use into another source of interaction cost. A web agent therefore needs persistent strategic context for the user's goal and on-demand tactical guidance for the current page state.

Consider a household agent asked to clean an object in an unfamiliar room. Early in the episode, it needs a strategy: a task-level guideline for locating the object, examining its state, and choosing an appropriate receptacle. Later, when the agent encounters a closed container or an exhausted search, it needs a tactic: a step-level technique for that specific local obstacle. Showing all tactics at every step mixes these two roles, consumes prompt budget, and gives the policy no reason to distinguish general task structure from local operational advice. Strategies should frame the entire episode, while tactics should be loaded only when the current observation calls for them.

Figure 1 illustrates the resulting progressive-disclosure framework for agentic skills, but most skill-augmented methods do not follow this design. Prior methods typically inject skills as a flat pool. Even hierarchical or graph-structured banks expose skills through a fixed delivery rule rather than letting the policy decide what to load given the current state [4, 38]. Under progressive disclosure, the agent instead receives persistent strategic context and a compact menu of associated tactics, then loads the detailed tactic only when the current state makes that detail useful.

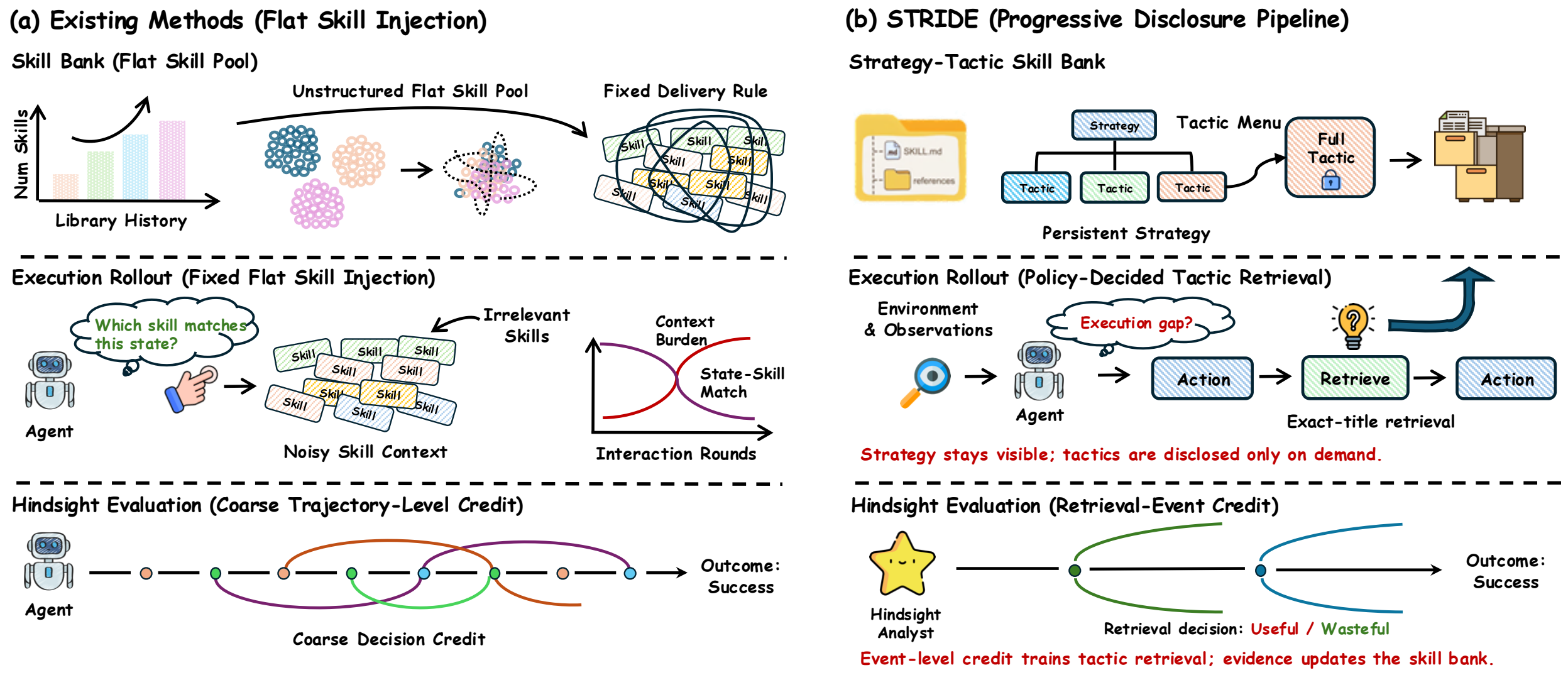

**Figure 1: Comparison of skill delivery paradigms. (a) Flat skill injection exposes the same skill pool across decisions and updates skill quality from trajectory-level outcomes. (b) STRIDE keeps the strategy visible, discloses tactics through a compact menu, and records each retrieval event as evidence for policy feedback and skill-bank updates.**

When tactic retrieval becomes an action, episode-level success is too coarse to credit individual retrieval decisions. A single outcome reward reveals only whether the episode succeeded, not whether a particular retrieval was timely or wasteful. Skill1 [22] decomposes rollout outcomes across lifecycle stages, and SAPO [50] estimates marginal utility through paired rollouts, but both still judge skills mainly from rollout-level evidence. STRIDE instead assigns feedback directly to each retrieval decision. This event-level signal teaches the policy when to retrieve, while the same retrieval traces give the strategy-tactic skill bank evidence for creating, updating, or pruning individual strategies and tactics.

We propose **STRIDE** (**S**trategy-to-**T**actics **R**etrieval-guided sk**I**ll **D**isclosure and **E**volution), a framework that embeds progressive skill disclosure within an RL training loop. STRIDE keeps the selected strategy visible, shows a compact tactic menu, and loads a full tactic only after the agent issues an explicit retrieval action, `<retrieve>`*exact tactic title*`</retrieve>`. This action creates a traceable decision point. The Tactician learns when to retrieve or act directly, the Analyst assigns credit to each retrieval decision in hindsight, and the Strategist uses retrieval traces to maintain the skill bank.

The main contributions are:

(1) **Strategy-to-tactics skill disclosure.** STRIDE separates skills by execution role: a task-level strategy stays visible throughout the episode, while step-level tactics appear as a compact menu and are loaded only through explicit retrieval. Skill exposure therefore depends on the current state rather than a fixed delivery rule.
(2) **Retrieval-event credit for learning when to use tactics.** STRIDE treats each tactic retrieval as an event-level credit unit. The Analyst assigns feedback based on whether the retrieval is needed, whether it matches the current state, and whether it is supported by subsequent rollout evidence, giving the policy a direct signal for when to retrieve.
(3) **A closed loop between policy learning and skill-bank evolution.** STRIDE feeds retrieval-conditioned evidence back into the skill bank to maintain and reorganize its strategies and tactics. This links tactic use, process feedback, and skill-bank evolution in one training loop. On ALFWorld, STRIDE achieves 92.7% average success with Qwen3-4B and 94.9% with Qwen3-8B, and it also improves over the strongest reproduced skill-augmented baseline on WebShop.

## 2 Related Work

### 2.1 Skill Libraries and Experience Reuse

Skill libraries convert prior trajectories into reusable action knowledge, so an agent can benefit from experience beyond the current episode. Early skill-augmented RL systems focus on producing and reusing such skills. SkillRL [36] recursively expands a skill set during RL, while INSPO [52] studies instruction-policy co-evolution rather than a persistent strategy-tactic bank. Skill1 [22] unifies skill evolution with policy optimization across skill lifecycle stages, and SAPO [50] estimates skill utility through paired rollouts. Whereas SkillRL retrieves task-specific skills by semantic similarity and conditions its policy on the retrieved content, STRIDE lets the policy explicitly request tactics under a persistent strategy and assigns feedback to each retrieval event. This line then expands from skill creation to skill-bank maintenance. Existing work studies how the bank should be revised, structured, or generalized [10, 12, 21, 45, 46], and the growing number of such systems has motivated benchmarking and systematization efforts [13]. Another branch examines what happens after skills are collected: skills can be internalized into the policy, curated for self-evolution, or retrieved conditionally rather than injected indiscriminately [5, 16, 17, 25, 48, 54]. ProactAgent [5] is closest on proactive retrieval: it learns when to ask for memory or skill content from a structured experience base. STRIDE differs by making the retrieved unit a tactic under a visible strategy, assigning PRM feedback to each retrieval event, and using the same event trace to update the linked strategy-tactic bank. Together, these systems show that skill banks become more effective when creation, selection, and maintenance are treated as connected problems. STRIDE contributes a disclosure mechanism that couples these operations at the retrieval-event level.

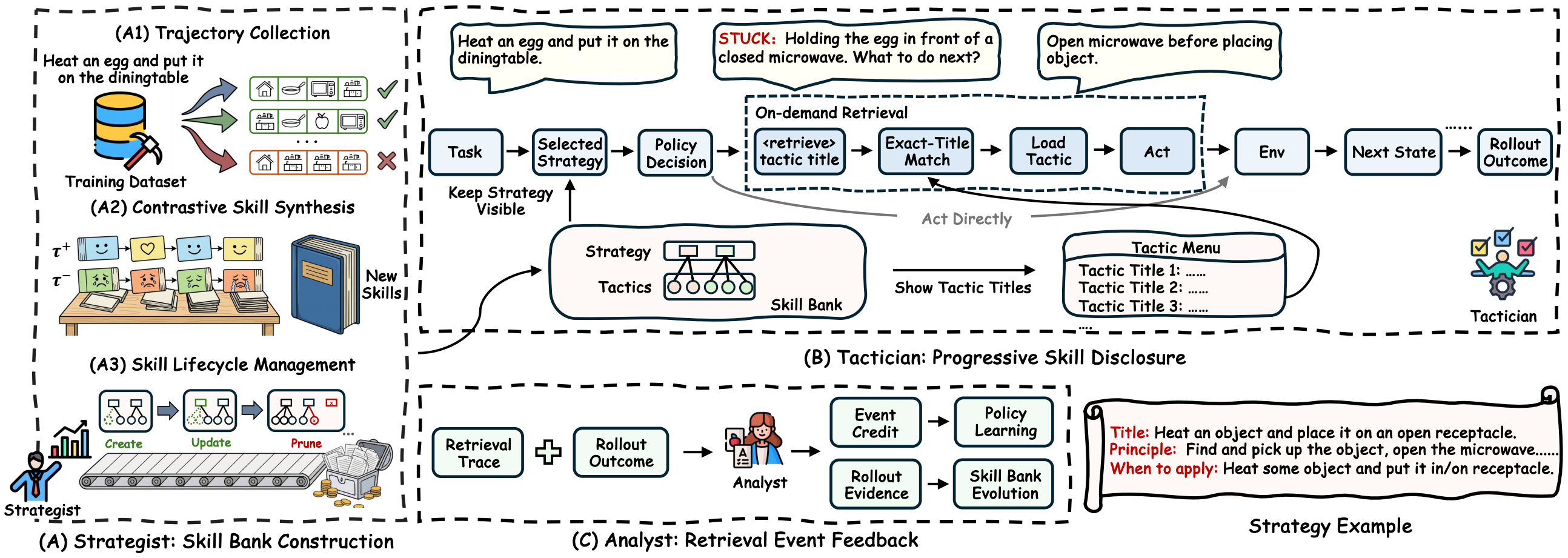

**Figure 2: Overview of STRIDE. The Tactician acts with a visible strategy and retrieves tactics only when the current state needs local guidance. The Analyst assigns event-level feedback to retrieval decisions, and the Strategist uses retrieval traces and rollout outcomes to create, update, or prune the strategy-tactic skill bank.**

Closest to STRIDE, D2Skill [28] maintains a dynamic dual-granularity skill bank that separates task-level and step-level knowledge. This design recognizes that reusable experience exists at different temporal scopes, but the two levels remain weakly coupled: a task-level skill does not explicitly define the step-level tools that should become available under it, and skill use is not treated as a policy decision with its own credit signal. STRIDE keeps the same temporal distinction but turns it into a disclosure interface. A selected strategy defines the visible tactic menu, and each tactic retrieval becomes both a learnable policy event and evidence for updating the linked strategy-tactic pair.

## 2.2 Memory-Augmented Reinforcement Learning

Memory-augmented RL studies the broader question of how persistent experience should be built, revised, and used during training. EvolveR [34], MemRL [47], and SimpleMem [15] address experience lifecycles, RL-updated memory banks, and efficient long-term memory, respectively. Recent systems extend this direction by making memory trainable or generative [37, 44, 53], refining memories from self-judged outcomes [6, 18], densifying memory-editing rewards [42], or sharing experience across tasks [27, 43]. These works establish memory as a learnable component of agent training, but memory use and policy optimization often remain separated: the system revises what is stored, while the policy receives memory through a retrieval or injection rule that is not itself the main credit-assignment target.

Recent agent-RL work moves toward dense feedback by assigning credit to intermediate steps, progress signals, turns, or hindsight events [7, 9, 26, 29, 32, 33]. STRIDE connects this dense-feedback view with skill memory. Instead of treating memory retrieval as external context preparation, it makes tactic retrieval an explicit policy action, assigns feedback to that event, and reuses the same trace for skill-bank evolution. The resulting loop links three operations that are often handled separately: deciding when to use memory, learning from that decision, and updating the memory that future decisions will expose.

# 3 Method

As shown in Figure 2, STRIDE organizes skills into a strategy-tactic skill bank and assigns three roles to manage its disclosure and evolution. The *Tactician* is the acting policy: it observes the current environment state, reads the persistent strategy, and decides whether to execute an environment action or retrieve a tactic. The *Analyst* is a hindsight process evaluator that assigns credit, via a process reward model (PRM), to the retrieval decisions made during the rollout. The *Strategist* updates the skill bank from those retrieval traces and the final rollout outcome. The three roles keep action selection, event-level credit, and memory maintenance coupled through the same retrieval events while separating their responsibilities. The method therefore has three linked parts: a disclosure interface that decides what information is visible, a retrieval action that lets the policy request missing local guidance, and an event-level feedback loop that trains retrieval behavior while updating the bank.

## 3.1 Preliminaries

An agent interacts with an environment $\mathcal{E}$ and produces a rollout $\tau = (o_0, a_0, \ldots, o_T, a_T)$ for task instruction $d$. At timestep $t$, policy $\pi_\theta(a_t \mid o_{\le t}, d, c)$ selects an action given the interaction history and additional context $c$. In a standard setting, $a_t$ belongs to the environment action space $\mathcal{A}_{\text{env}}$ and the rollout yields an environment return $R_i^{\text{env}}$. We track the binary success indicator $Y_i \in \{0, 1\}$ separately for success-rate reporting. STRIDE changes the context $c$ and the action space: $c$ contains the selected strategy and tactic-title menu, while $a_t$ may also be a retrieval command that loads a tactic into the history. The training objective remains task success, but the learning signal is enriched by feedback on these retrieval events. This formulation preserves the underlying RL problem while making skill use observable as part of the trajectory.

**Group Relative Policy Optimization (GRPO).** Following Shao et al. [20] and related REINFORCE-style LLM optimization work [1], we optimize the policy by sampling a group of $G$ rollouts $\{\tau_i\}_{i=1}^{G}$ per task, setting each rollout return to the environment return $R_i^{\text{env}}$, and computing the group-normalized advantage

$$A_i = \frac{R_i^{\text{env}} - \text{mean}(\{R_j^{\text{env}}\}_{j=1}^{G})}{\text{std}(\{R_j^{\text{env}}\}_{j=1}^{G})}, \tag{1}$$

then maximizing a clipped surrogate objective [19] with KL regularization:

$$\max_\theta \ \mathcal{J}_{\text{GRPO}}(\theta) = \mathbb{E}\left[\frac{1}{G}\sum_{i=1}^{G} \min\Big(\rho_i A_i, \text{clip}(\rho_i, 1-\epsilon, 1+\epsilon) A_i\Big) - \beta D_{\text{KL}}\right] \tag{2}$$

where $\rho_i$ is the policy probability ratio, $\epsilon$ is the clipping range, and $\beta$ controls the KL penalty. STRIDE keeps this policy-optimization backbone but changes how the advantage is constructed: retrieval events receive Analyst feedback and are folded into the decision-level advantage, so the policy learns not only which environment action to take but also when tactic disclosure is worth using.

## 3.2 Progressive Skill Disclosure

STRIDE separates skill disclosure by temporal scope. A strategy remains visible because it encodes the task-level order of operations across many steps. A tactic is loaded on demand when the agent faces a local obstacle, such as deciding whether to check likely receptacles before expanding the search. This distinction gives the two layers different delivery modes. The skill bank $\mathcal{S}$ consists of a strategy set $\mathcal{S}_{\text{stra}}$, and each strategy $T \in \mathcal{S}_{\text{stra}}$ has an attached tactic set $\mathcal{S}_{\text{tac}}(T)$. The strategic layer is selected before the rollout to define the global context; the tactical layer remains latent until a state requires more specific guidance.

*Task-Level Guidelines (Strategic Layer).* In STRIDE, task-level skills are *strategies*: reusable guidelines that frame how to solve all instances of a task category. These guidelines are retrieved once at rollout start and injected into the prompt to establish strategic context before any environment action is taken. Following the UCB-augmented retrieval score used in RetroAgent [49] for skill selection, STRIDE scores each candidate strategy by combining task similarity, historical utility, and an exploration bonus:

$$T^* = \arg\max_{T \in \mathcal{S}_{\text{stra}}} r(T), \tag{3}$$

$$r(T) = \alpha \widetilde{\text{sim}}(e_d, e_T) + (1-\alpha)(u^T + b(T)), \tag{4}$$

$$b(T) = \eta \sqrt{\frac{\log(1 + N_{\text{all}})}{1 + N_r^T}}. \tag{5}$$

Here $e_d$ and $e_T$ are embeddings of the task instruction and strategy, $u^T$ is the strategy utility, $N_r^T$ is its retrieval count, and $N_{\text{all}}$ is the total retrieval count over strategy-layer skills. The similarity term selects a strategy relevant to the instruction, the utility term favors strategies that have supported successful rollouts, and the bonus prevents the bank from collapsing onto a small set of early high-utility strategies. The highest-scoring strategy $T^*$ stays visible throughout the episode and anchors the tactic menu used in later steps. In the implementation, the prompt can display the top $K$ strategy entries with up to $L$ tactic titles under each entry, matching the parameter table in Appendix A. Strategy selection is therefore a context-setting step rather than a repeated retrieval decision.

*Step-Level Techniques (Tactical Layer).* In STRIDE, step-level skills are *tactics*: concrete operational methods for specific local states. Each tactic records when it applies and what action pattern it recommends, but this detailed content is not injected by default. Each tactic belongs to exactly one strategy's tactic set $\mathcal{S}_{\text{tac}}(T)$. For each visible strategy entry, only its tactic titles appear in the menu. Full tactic content stays in the bank until the agent explicitly requests it. This design keeps the context compact while still giving the policy a visible set of local tools it can request.

*Index-then-Detail Menu.* STRIDE implements this disclosure with a lightweight menu, following the index-then-detail principle behind Anthropic's agent-skill design [2]. The agent loads a tactic by emitting <retrieve>*exact tactic title*</retrieve>, matching against visible titles in $\mathcal{S}_{\text{tac}}(T^*)$ rather than searching the whole bank. Unlike strategy selection, which is resolved automatically by $r(T)$, tactic retrieval is a policy decision learned during training. Exact-title retrieval also turns the menu into an auditable interface: every loaded tactic can be matched to the state in which it was requested and to the later outcome of the rollout. This interface keeps the prompt small before retrieval and creates a precise event boundary after retrieval.

## 3.3 Strategy-Contextualized Tactic Retrieval

Given a selected strategy, the Tactician decides whether the visible strategy and tactic titles suffice for the current state, or whether a local obstacle requires loading a concrete tactic. This decision is different from ordinary retrieval-augmented prompting: retrieval is not a preprocessing step outside the policy, but an action that competes with environment actions at each step. The policy therefore controls both task execution and the timing of skill disclosure.

*Extended Action Space.* We extend the agent's action space with an explicit retrieval action:

$$\mathcal{A} = \mathcal{A}_{\text{env}} \cup \{\texttt{<retrieve>}q\texttt{</retrieve>} \mid q \in Q\} \tag{6}$$

where $Q$ is the set of tactic titles shown in the visible strategy menus. Emitting a valid retrieval action loads the full content of the matched tactic into the agent's context. Invalid titles return no tactic and are recorded as failed retrieval attempts. A retrieval action does not advance the environment, so a useful retrieval must justify the extra decision step by improving the next environment action or avoiding a longer detour. This cost makes retrieval selective: the policy benefits from a tactic only when the retrieved content changes the subsequent action choice enough to offset the extra step.

*Learning Retrieval through RL..* Rather than learning retrieval through separate supervised fine-tuning or a fixed retrieval heuristic [34, 36], STRIDE trains the policy to choose between acting directly and loading a tactic from the visible menu. The policy therefore learns two coupled behaviors: recognizing states where the strategy is sufficient, and recognizing states where the title of a tactic indicates missing local guidance. This coupling targets the two retrieval failure modes directly: over-retrieval wastes steps and context, while under-retrieval leaves local bottlenecks unsolved.

## 3.4 Retrieval-Event Feedback and Skill Evolution

Once retrieval is part of the action space, each retrieval event becomes both a credit-assignment target for policy learning and a trace for skill-bank updates. STRIDE uses the same event record for both purposes, but separates the learning signal from the bank-maintenance rule: the Analyst scores whether a retrieval was useful in context, while the Strategist updates skill statistics over repeated use. This separation lets the policy learn from fine-grained retrieval quality without making every bank update depend on a single PRM judgment.

*Analyst: Hindsight Process Evaluation.* After rollout completion, the Analyst revisits each retrieval event in its local rollout context. Following process supervision and agent process-reward modeling [14, 31, 35], it uses a PRM to assign a scalar credit $r_t$ to the retrieval decision at step $t$. The score reflects three local questions: whether the current state required extra guidance, whether the requested tactic matched that state, and whether the subsequent action used the retrieved guidance. This converts retrieval from a hidden prompt-construction choice into a supervised decision point inside the rollout. The evaluation is hindsight-based because the usefulness of a tactic is often visible only after the agent acts with or ignores the retrieved guidance.

*Token-Level Shaped Reward.* For each valid retrieval event at decision step $t$ with Analyst score $r_t$, STRIDE adds a shaped reward to the last token of the retrieval response:

$$\hat{R}_{i,k} = \begin{cases} R_i^{\text{env}} + \lambda r_t, & k = \ell(t), \\ R_i^{\text{env}}, & \text{otherwise}, \end{cases} \tag{7}$$

where $k$ indexes token positions and $\ell(t)$ denotes the last token position of the retrieval emitted at step $t$. The shaped return $\hat{R}_{i,k}$ is group-normalized to obtain a token-level advantage $A_{i,k}$, and the clipped GRPO objective uses the corresponding per-token ratio $\rho_{i,k}$. The hyperparameter $\lambda$ controls the weight of process feedback relative to the outcome reward. Assigning the bonus to the retrieval response localizes the credit to the decision that requested a tactic, rather than raising the return of every action in the trajectory. The environment return still trains overall task completion, while the shaped term gives retrieval decisions their own learning signal.

*Retrieval-Conditioned Utility Updates.* In parallel with PRM feedback, retrieval traces and the binary success indicator $Y_i$ update utility statistics in the skill bank. For every retrieved tactic $s$, the bank tracks retrieval count $N_r^s$, effective count $N_e^s$, a sliding outcome history $H^s$, and an EMA utility $u^s$. Given the rollout success value $v = Y_i$, the utility update with EMA rate $\gamma$ is

$$u^s \leftarrow (1-\gamma)u^s + \gamma v. \tag{8}$$

The effective count $N_e^s$ increments when $s$ is retrieved in a successful episode ($v = 1$), and recent outcomes are stored in $H^s$ for harmful-skill detection. These statistics separate consistently useful tactics from tactics that are retrieved often but rarely lead to success. The PRM score captures event quality, while the utility statistics capture longer-run association with successful task completion. Using both signals prevents a tactic from being kept solely because it is frequently requested.

*Strategist: Lifecycle Management of Skill Bank.* The Strategist turns retrieval traces into lifecycle decisions for the skill bank. Each event records the selected strategy, the retrieved tactic, the post-retrieval action, and the rollout outcome. The lifecycle rules address three bank-maintenance cases: missing tactics are created, stale or misaligned descriptions are updated, and low-value entries are pruned from the visible menu.

**Create.** When a rollout fails, the Strategist compares it with a paired successful rollout and asks an external LLM to distill a reusable tactic from their contrast. The new tactic is mounted under the candidate strategy that frames the failed attempt. If no candidate strategy covers the scope, the same contrastive pair is distilled into a new strategy.

**Update.** As tactics accumulate under a strategy, or when a strategy remains weak despite repeated recent use, the Strategist rewrites the strategy from its associated tactics so the menu stays aligned with the obstacles the agent actually encounters.

**Prune.** A tactic is removed when it is rarely retrieved (low $N_r^s$) or when $H^s$ shows it to be persistently ineffective or harmful over sustained use. A strategy is removed when it retains no useful tactics, or when it remains low-utility despite earlier rewrite attempts. When the bank still exceeds capacity $N_{\max}$ after these removals, remaining tactics are ranked by

$$V(s) = \frac{N_e^s}{N_r^s} \cdot u^s, \tag{9}$$

and the lowest-scoring, non-recent entries are trimmed to keep stale tactics out of the tactic menu.

## 3.5 Training Procedure

STRIDE uses a two-phase training curriculum to separate format acquisition from retrieval selectivity. This curriculum addresses the cold-start problem created by a new action type: before the policy can learn when retrieval is useful, it must first learn to produce a valid retrieval command. Early training therefore makes the retrieval action available to the policy. Later training makes retrieval useful only when it addresses a real obstacle.

**Table 1: Success rate (%) on ALFWorld across Seen and Unseen task splits and three subtask types, using Qwen3-4B-Instruct and Qwen3-8B backbones. Within each Qwen backbone group, the best result in each column is in bold; the second-best is underlined.**

| | | Seen | | | | Unseen | | | | |
|---|---|---|---|---|---|---|---|---|---|---|
| | Method | Pick&Place | Transform | Examine | Avg | Pick&Place | Transform | Examine | Avg | Overall |
| *Proprietary* | Sonnet 4.5 | 89.8 | 73.5 | 96.2 | 82.5 | 73.2 | 72.0 | 100.0 | 76.1 | 79.2 |
| | **Qwen3-4B-Instruct** | | | | | | | | | |
| *Baseline* | ReAct | 54.7 | 28.1 | 28.6 | 37.1 | 30.8 | 28.9 | 66.7 | 32.8 | 35.0 |
| *w/ Evolve* | ReMe | 60.5 | 44.6 | 79.5 | 54.5 | 45.5 | 58.2 | 46.3 | 52.7 | 53.6 |
| | MemRL | 52.0 | 30.9 | 56.4 | 42.1 | 41.5 | 44.4 | 55.6 | 45.0 | 43.6 |
| *w/ RL* | GRPO | **93.8** | 62.3 | 53.8 | 74.8 | 86.2 | 72.0 | 68.5 | 75.9 | 75.3 |
| | MemRL (+RL) | 89.8 | 75.8 | 71.8 | 81.2 | 78.9 | 76.6 | 94.4 | 79.1 | 80.2 |
| | EvolveR | 70.6 | 85.2 | 61.5 | 76.7 | 69.9 | 77.9 | 94.4 | 77.9 | 77.3 |
| | INSPO | 78.0 | 77.6 | 61.5 | 76.4 | 62.9 | 87.0 | 83.3 | 79.6 | 78.0 |
| | SkillRL | 89.8 | 85.9 | 61.5 | 85.7 | 69.9 | 86.9 | 88.9 | 82.1 | 83.9 |
| | ReSkill | 91.6 | 88.7 | 76.9 | 90.0 | 87.4 | 89.6 | **96.3** | 89.6 | 89.8 |
| | **Ours** | 92.8 | **92.7** | **92.3** | **92.8** | **90.2** | **92.0** | 94.4 | **92.5** | **92.7** |
| | **Qwen3-8B** | | | | | | | | | |
| *Baseline* | ReAct | 73.4 | 44.8 | 64.3 | 59.3 | 69.2 | 65.0 | 54.7 | 64.9 | 62.0 |
| *w/ Evolve* | ReMe | 79.1 | 58.3 | 48.7 | 66.2 | 53.7 | 76.4 | 48.1 | 65.7 | 65.9 |
| | MemRL | 64.4 | 37.7 | 61.5 | 51.2 | 57.7 | 58.7 | 57.4 | 58.2 | 54.9 |
| *w/ RL* | GRPO | 93.8 | 72.1 | 66.7 | 80.7 | 82.1 | 81.8 | 79.6 | 81.6 | 81.1 |
| | MemRL (+RL) | 86.4 | 78.5 | 84.6 | 83.1 | 82.1 | 83.3 | 75.9 | 81.6 | 82.3 |
| | EvolveR | 88.7 | 79.8 | 74.4 | 83.3 | 70.7 | 89.2 | 83.3 | 82.8 | 83.1 |
| | INSPO | 90.4 | 79.2 | 71.8 | 83.2 | 88.1 | 89.2 | 81.5 | 87.8 | 85.5 |
| | SkillRL | 95.5 | 82.8 | 87.2 | 89.0 | 82.9 | 78.3 | 94.4 | 82.6 | 85.8 |
| | ReSkill | 96.5 | 81.4 | **94.9** | 90.2 | **99.3** | 91.2 | **100.0** | **95.3** | 92.7 |
| | **Ours** | **98.3** | **94.1** | 92.3 | **95.7** | 97.6 | **98.7** | 66.7 | 94.0 | **94.9** |

**Phase 1: Retrieval Behavior Bootstrapping.** For the first $T_{\text{warmup}}$ steps, each syntactically valid retrieval receives a rollout-level bonus. This dense reward teaches the policy to emit the <retrieve>...</retrieve> format and use the index-then-detail menu described in Section 3.2. At this stage, the objective is not to judge whether every retrieval is necessary. It is to make the retrieval action part of the policy's reachable behavior.

**Phase 2: Necessity-Conditioned Retrieval.** After $T_{\text{warmup}}$, the format bonus decays, per-event Analyst credit becomes the primary retrieval signal, and retrievals beyond a rollout-level cap are penalized. The policy is then rewarded for retrieving tactics that match the current state and discouraged from treating the menu as a generic context-expansion mechanism.

## 4 Experiments

The experiments evaluate STRIDE along a single evidence chain. The main comparison tests whether strategy-to-tactics disclosure improves task success across embodied and web interaction. The skill-bank analysis then asks whether retrieval traces produce a more useful bank over training, and the ablations isolate which parts of the interface and learning signal matter. Finally, the dynamics and case studies explain how retrieval behavior changes and when tactics provide information beyond the persistent strategy.

### 4.1 Experimental Setup

**Environments.** We evaluate on two interactive benchmarks with different decision structures. *ALFWorld* [24] is a text-based embodied environment where the agent must locate, transform, and examine household objects through long action sequences with strict precondition ordering. *WebShop* [40] simulates online shopping, requiring the agent to navigate product pages, compare attributes, and purchase an item matching a user instruction. ALFWorld tests long-horizon precondition management, where strategies should preserve global order and tactics should resolve local search or manipulation bottlenecks. WebShop tests whether the same retrieval interface transfers to a web navigation setting with shorter horizons and attribute comparison. On ALFWorld, we report the success rate for Pick&Place, Transform, and Examine tasks on both Seen and Unseen splits, together with split averages and an overall average. We test Qwen3-4B-Instruct-2507 (abbreviated as Qwen3-4B) and Qwen3-8B [3] backbones. On WebShop, we report the task score and success rate with Qwen2.5-7B-Instruct. D2Skill reports near-zero

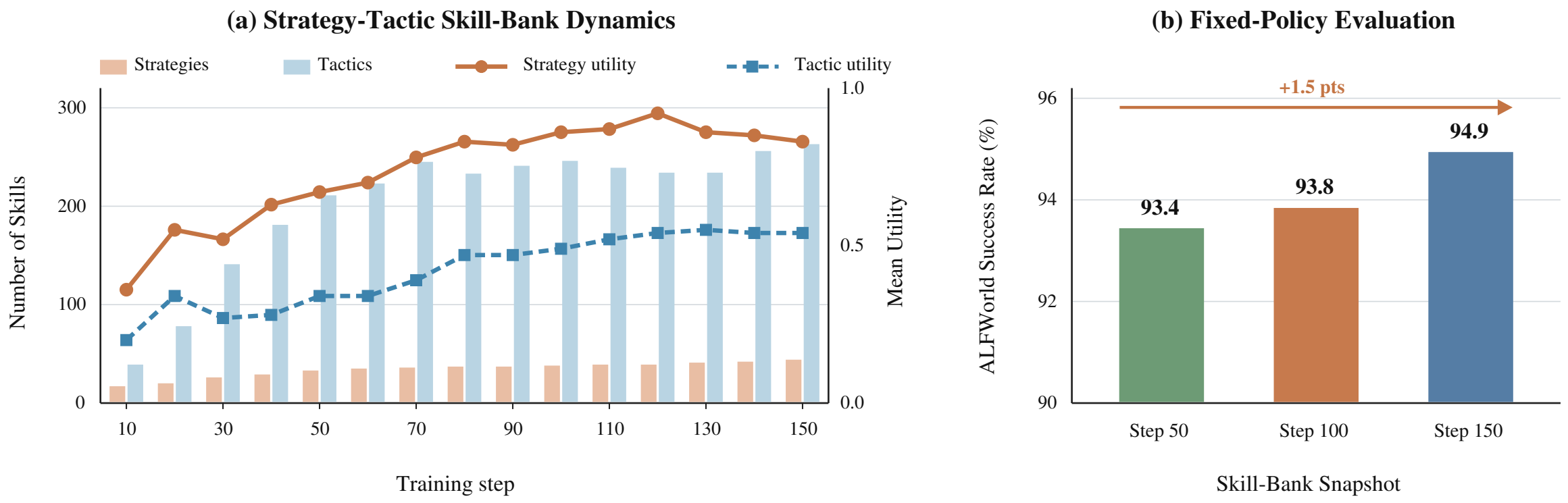

**Figure 3: Skill-bank evolution on ALFWorld. (a) Bank growth occurs mainly in the tactic layer, while the number of strategies changes little and utility rises early before stabilizing. (b) Under a fixed final policy, success increases from 93.4% with the Step-50 bank to 94.9% with the Step-150 bank.**

WebShop metrics for its Qwen3-4B checkpoint without supervised fine-tuning [28]. We therefore use Qwen2.5-7B-Instruct to keep the RL-only training setting consistent and avoid adding a supervised fine-tuning stage that would confound the skill-retrieval comparison. Figures 3, 4, and 5 summarize skill-bank evolution, training dynamics, and representative retrieval behavior, respectively.

**Auxiliary model.** Sonnet 4.5 serves as the Analyst for post-rollout retrieval scoring and as the Strategist for skill creation and revision (Table 6 in the appendix). The acting policy learns through RL with environment rewards and retrieval-event feedback, rather than supervised imitation of Sonnet-generated action trajectories.

**Baselines.** On ALFWorld, we include Sonnet 4.5 as a proprietary reference and compare against ReAct prompting [41], the non-RL memory evolution method ReMe [6], and several policy-training or skill/memory baselines: MemRL without RL, EvolveR, INSPO's instruction-policy co-evolution, SkillRL, ReSkill, GRPO, and MemRL (+RL). Table 2 reports the WebShop comparison under the Qwen2.5-7B-Instruct setting against ReAct, Reflection, MemRL, GRPO, EvolveR, Mem0+GRPO, SimpleMem+GRPO, and SkillRL. These baselines cover prompting-only agents, memory evolution without RL, outcome-reward RL, and skill-augmented RL, so the comparison separates the strategy-tactic interface from generic training gains. Appendix C explains why ProactAgent is not included as a reported baseline in this version.

## 4.2 Main Results

Table 1 tests the primary effectiveness claim: whether progressive skill disclosure improves long-horizon agent execution. STRIDE achieves the highest overall ALFWorld success rate at both model scales. It reaches 92.7% with Qwen3-4B, outperforming ReSkill by 2.9 points, and 94.9% with Qwen3-8B, a 2.2-point gain. The gains are concentrated on Pick&Place and Transform, the two subtasks that require longer action chains and stricter precondition ordering. These results align with the method's design: persistent strategies preserve the order of the task, while tactics resolve local search and manipulation bottlenecks. Both RL-trained backbones also exceed the Sonnet 4.5 reference (79.2% overall; Table 1), showing that the learned execution policy can surpass the auxiliary model's measured task performance.

**Table 2: WebShop results under the Qwen2.5-7B-Instruct setting. We report task score and success rate (Succ.); best results are in bold.**

| | Method | Score | Succ. |
|---|---|---|---|
| *Baseline* | Qwen2.5-7B-Instruct | 26.4 | 7.8 |
| *w/ Prompt* | ReAct | 46.2 | 19.5 |
| | Reflection | 58.1 | 28.8 |
| *w/ Evolve* | MemRL | 29.5 | 9.2 |
| *w/ RL* | GRPO | 79.3 | 66.1 |
| | EvolveR | 42.5 | 17.3 |
| | Mem0+GRPO | 58.1 | 37.5 |
| | SimpleMem+GRPO | 67.8 | 46.9 |
| | SkillRL | 85.2 | 72.7 |
| | **Ours** | **89.7** | **74.6** |

The improvement is not uniform across every cell. On Qwen3-8B Unseen Examine, STRIDE underperforms ReSkill, even though it remains strongest overall. This boundary is consistent with the role of STRIDE's hierarchy. The method reduces search and ordering errors most directly, while an Examine task can depend more heavily on direct object-state inspection after the target is reached.

The WebShop results in Table 2 test whether the same retrieval mechanism transfers beyond embodied interaction. STRIDE improves over SkillRL by 4.5 score points and 1.9 success-rate percentage points. WebShop requires the agent to keep a user goal stable while inspecting page-specific product evidence, so the result supports the same strategy-to-tactics principle in a web setting.

## 4.3 Skill-Bank Evolution

The bank develops in three phases: strategies and tactics both grow early on, tactics keep growing after strategies plateau, and pruning later offsets new creation so bank size holds steady while utility remains elevated. This pattern matches the intended hierarchy. A small set of strategies captures recurring task-level plans, while tactics continue to specialize around local obstacles encountered under those strategies.

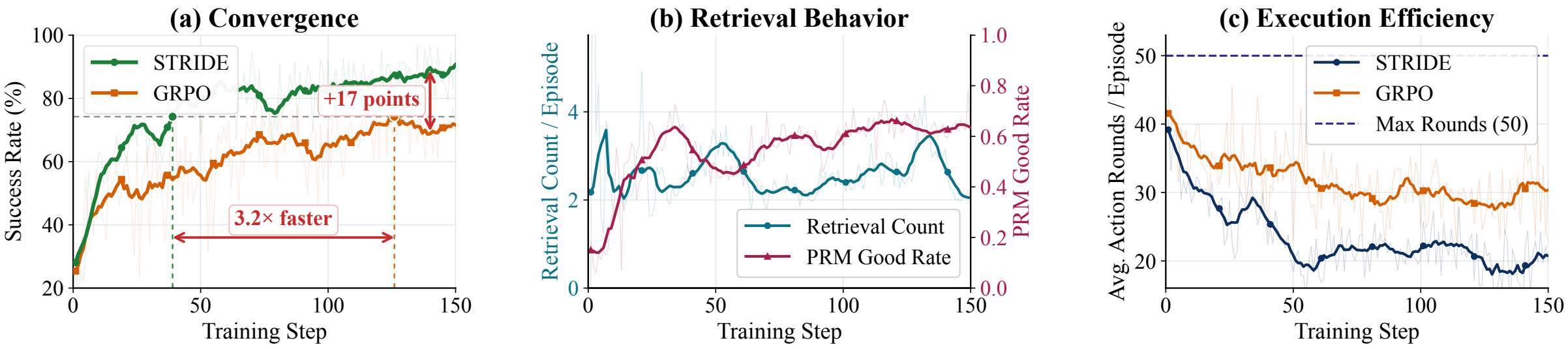

**Figure 4: Training dynamics on ALFWorld with Qwen3-4B. (a) STRIDE reaches GRPO's peak success rate in 1/3.2 as many training steps and maintains an approximately 17-point advantage near the end of training. (b) Retrieval frequency decreases after warm-up as the Analyst GOOD-label rate rises. (c) STRIDE uses fewer action rounds than GRPO.**

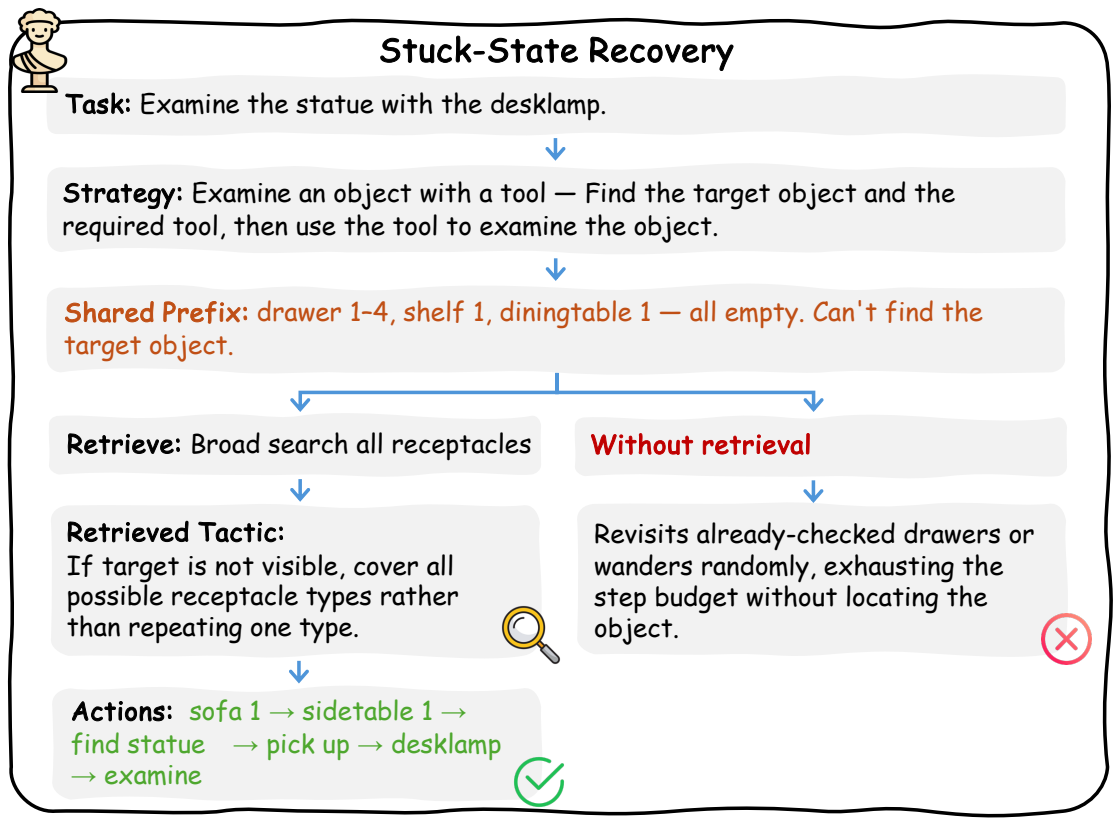

**Figure 5: A stuck-state recovery example. The retrieved tactic redirects the agent from repeating already checked locations to broadening the search over remaining receptacles.**

To measure bank-snapshot performance under a fixed actor, Figure 3(b) fixes the final Qwen3-8B policy and tests it against bank snapshots from Steps 50, 100, and 150. Success rises monotonically from 93.4% to 93.8% to 94.9%. This fixed-policy protocol shows that later bank snapshots are more effective for the final policy to consume; it does not by itself isolate a standalone bank-quality improvement independent of that policy.

## 4.4 Ablation Study

Table 3 separates two sources of evidence on ALFWorld with Qwen3-4B. The inference-time ablations remove visible skill components from a trained STRIDE policy, so they measure which information the policy uses at test time. The training-time ablations remove learning signals or retrieval capacity during optimization, so they measure which mechanism is needed to acquire the behavior.

The inference-time ablations reveal an asymmetric dependence. Removing the strategy reduces success by 19.8 points and adds 8.8 rounds, whereas removing tactics reduces success by 3.6 points and adds 3.5 rounds. The persistent strategy is therefore the main source of task-level reliability: without it, the agent loses the global order that constrains long-horizon household tasks. Tactics provide a smaller but consistent gain by shortening local detours once the task-level plan is already in place.

**Table 3: Ablation study on ALFWorld with Qwen3-4B. SR is success rate (%); Rounds is the average number of action steps per episode.**

| Variant | SR (%) ↑ | Rounds ↓ |
|---|---|---|
| *Inference-Time Skill Ablations* | | |
| STRIDE | **92.7** | **17.3** |
| w/o Tactic | 89.1 | 20.8 |
| w/o Strategy | 72.9 | 26.1 |
| *Training-Time Component Ablations* | | |
| w/o Analyst | 85.0 | 24.0 |
| w/o Analyst + w/o Tactic | 84.3 | 20.0 |
| GRPO | 75.3 | 25.9 |
| ReAct (no training) | 35.0 | 42.3 |

Training without the Analyst lowers success from 92.7% to 85.0% and increases the average trajectory from 17.3 to 24.0 rounds. Removing both Analyst feedback and tactic retrieval yields a similar success rate of 84.3%. This result separates access from credit: making tactics available is not enough unless the policy receives a signal for when retrieval is warranted. Relative to GRPO, the complete system improves success by 17.4 points and saves 8.6 rounds, indicating that the hierarchy and event-level training signal work together.

## 4.5 Training Dynamics

The training curves explain how STRIDE converts retrieval into an increasingly selective policy behavior. We connect three measurements: how quickly success improves, how retrieval behavior changes, and whether the resulting policy executes tasks with fewer environment actions.

**Training-step convergence.** Figure 4(a) shows that STRIDE reaches GRPO's peak success rate in 1/3.2 as many training steps and retains an approximately 17-point advantage near the end of training. This comparison is based on logged training steps, not wall-clock time or total token cost. The trend is consistent with the Analyst ablation in Table 3: retrieval-event feedback adds a decision-level signal beyond the rollout outcome reward.

**Table 4: Trigger types for tactic retrieval in validation rollouts. The first two act before failure; the third responds after the search becomes stuck.**

| Trigger Type | Purpose | Share |
|---|---|---|
| Procedural Error Prevention | Preserve task order | 47.5% |
| Prior-Guided Localization | Prioritize likely locations | 27.1% |
| Stuck-State Recovery | Broaden exhausted search | 25.4% |

**Retrieval behavior evolution.** Retrieval frequency in Figure 4(b) falls to roughly two events per episode by the end of training, while the Analyst GOOD-label rate rises from about 0.1 to about 0.8. This rate is the fraction of retrieval events labeled GOOD by the training Analyst prompt, not an independent human validation score. The two trends separate selective retrieval from indiscriminate skill use: the policy retrieves less often as training proceeds, but a larger fraction of those decisions match the local state under the Analyst rubric.

**Execution efficiency.** Figure 4(c) shows a persistent gap in environment action rounds after the early training stage. This is an environment-step measure rather than a wall-clock or total-token-cost measurement. Together with the ablation results, the curve indicates that strategies reduce task-level detours and tactics shorten recovery at specific bottlenecks.

## 4.6 Case Study: When Does Tactic Retrieval Help?

Figure 5 illustrates a representative stuck-state recovery case, where the strategy preserves the task goal but the retrieved tactic changes the local search decision. Table 4 annotates tactic retrievals in validation rollouts by the state that triggers them. Procedural error prevention occurs when the task is understood but an action dependency remains implicit. For example, *"locate target before cleaning"* preserves the required order. Stuck-state recovery appears only after the prioritized search has failed, when *"broaden search to all receptacles"* becomes the appropriate response.

The counts clarify what tactics contribute. Preventive retrievals account for 74.6% of all events, while 25.4% occur after the search is already stuck. Most tactics therefore resolve a pending action choice rather than repair a failure. The persistent strategy supplies task structure, and retrieval adds a state-specific constraint only when that structure alone does not determine the next action.

# 5 Conclusion

STRIDE starts from the observation that reusable agent knowledge serves different temporal roles: some knowledge should frame the whole episode, while other knowledge should be disclosed only at local bottlenecks. STRIDE operationalizes this distinction with a persistent strategy, an on-demand tactic menu, event-level feedback for retrieval, and retrieval traces for skill-bank evolution. It achieves 92.7% and 94.9% average success on ALFWorld with Qwen3-4B and Qwen3-8B, respectively, and improves over the strongest reproduced skill-augmented baseline on WebShop by 4.5 score points and 1.9 success-rate percentage points. Ablations show that the strategy provides the main source of task-level reliability, while tactics and retrieval feedback reduce environment action rounds. The results support a direct design principle for skill-augmented agents: reusable knowledge is most useful when its abstraction level and disclosure timing match the decisions it is meant to support.

# References

[1] Arash Ahmadian, Chris Cremer, Matthias Gallé, Marzieh Fadaee, Julia Kreutzer, Olivier Pietquin, Ahmet Üstün, and Sara Hooker. 2024. Back to Basics: Revisiting REINFORCE-Style Optimization for Learning from Human Feedback in LLMs. In *Proceedings of the 62nd Annual Meeting of the Association for Computational Linguistics (Volume 1: Long Papers)*. 12248–12267. doi:10.18653/v1/2024.acl-long.662

[2] Anthropic. 2025. Equipping Agents for the Real World with Agent Skills. https://www.anthropic.com/engineering/equipping-agents-for-the-real-world-with-agent-skills. Accessed: 2026-07-29.

[3] Jinze Bai, Shuai Bai, Yunfei Chu, Zeyu Cui, Kai Dang, Xiaodong Deng, Yang Fan, Wenbin Ge, Yu Han, Fei Huang, et al. 2023. Qwen Technical Report. arXiv:2309.16609 [cs.CL]

[4] Tong Bai, Zhenglin Wan, Pengfei Zhou, Xingrui Yu, Yang You, and Ivor W. Tsang. 2026. SkillDAG: Self-Evolving Typed Skill Graphs for LLM Skill Selection at Scale. arXiv:2606.03056 [cs.AI]

[5] Yuxuan Cai, Wei Li, Jie Zhou, Qin Chen, Xin Li, Bo Zhang, and Liang He. 2026. Ask Only When Needed: Proactive Retrieval from Memory and Skills for Experience-Driven Lifelong Agents. arXiv:2604.20572 [cs.CL]

[6] Zouying Cao, Jiaji Deng, Li Yu, Weikang Zhou, Zhaoyang Liu, Bolin Ding, and Hai Zhao. 2025. Remember Me, Refine Me: A Dynamic Procedural Memory Framework for Experience-Driven Agent Evolution. arXiv:2512.10696 [cs.AI]

[7] Yanjun Chen, Yirong Sun, Hanlin Wang, Jinghan Wang, Xinming Zhang, Xiaoyu Shen, Wenjie Li, and Wei Zhang. 2026. Exact Is Easier: Credit Assignment for Cooperative LLM Agents. arXiv:2603.06859 [cs.LG]

[8] Prateek Chhikara, Dev Khant, Saket Aryan, Taranjeet Singh, and Deshraj Yadav. 2025. Mem0: Building Production-Ready AI Agents with Scalable Long-Term Memory. arXiv:2504.19413 [cs.CL]

[9] Lang Feng, Zhenghai Xue, Tingcong Liu, and Bo An. 2025. Group-in-Group Policy Optimization for LLM Agent Training. arXiv:2505.10978 [cs.LG]

[10] Zelin He, Haotian Lin, Boran Han, Wei Zhu, Haoyang Fang, Bernie Wang, Xuan Zhu, Runze Li, and Matthew Reimherr. 2026. ReSkill: Reconciling Skill Creation with Policy Optimization in Agentic RL. arXiv:2606.01619 [cs.AI]

[11] Shixin Hong, Yu Liu, Zhi Li, Shaohui Li, and You He. 2024. Multi-Agent Collaborative Perception via Motion-Aware Robust Communication Network. In *Proceedings of the IEEE/CVF Conference on Computer Vision and Pattern Recognition (CVPR)*. 15301–15310. doi:10.1109/CVPR52733.2024.01449

[12] Xiaoyuan Li, Moxin Li, Keqin Bao, Yubo Ma, Wenjie Wang, Dayiheng Liu, and Fuli Feng. 2026. SkillGraph: Skill-Augmented Reinforcement Learning for Agents via Evolving Skill Graphs. arXiv:2605.12039 [cs.CL]

[13] Xiangyi Li, Yimin Liu, Wenbo Chen, Bingran You, Zonglin Di, Yifeng He, et al. 2026. SkillsBench: Benchmarking How Well Agent Skills Work Across Diverse Tasks. arXiv:2602.12670 [cs.AI]

[14] Hunter Lightman, Vineet Kosaraju, Yura Burda, Harri Edwards, Bowen Baker, Teddy Lee, Jan Leike, John Schulman, Ilya Sutskever, and Karl Cobbe. 2023. Let's Verify Step by Step. arXiv:2305.20050 [cs.LG]

[15] Jiaqi Liu, Yaofeng Su, Peng Xia, Siwei Han, Zeyu Zheng, Cihang Xie, Mingyu Ding, and Huaxiu Yao. 2026. SimpleMem: Efficient Lifelong Memory for LLM Agents. arXiv:2601.02553 [cs.AI]

[16] Zhengxi Lu, Zhiyuan Yao, Jinyang Wu, Chengcheng Han, Qi Gu, Xunliang Cai, Weiming Lu, Jun Xiao, Yueting Zhuang, and Yongliang Shen. 2026. SKILL0: In-Context Agentic Reinforcement Learning for Skill Internalization. arXiv:2604.02268 [cs.LG]

[17] Siru Ouyang, Jun Yan, Yanfei Chen, Rujun Han, Zifeng Wang, Bhavana Dalvi Mishra, Rui Meng, Chun-Liang Li, Yizhu Jiao, Kaiwen Zha, Maohao Shen, Vishy Tirumalashetty, George Lee, Jiawei Han, Tomas Pfister, and Chen-Yu Lee. 2026. SkillOS: Learning Skill Curation for Self-Evolving Agents. arXiv:2605.06614 [cs.AI]

[18] Siru Ouyang, Jun Yan, I-Hung Hsu, Yanfei Chen, Ke Jiang, Zifeng Wang, Rujun Han, Long T. Le, Samira Daruki, Xiangru Tang, Vishy Tirumalashetty, George Lee, Mahsan Rofouei, Hangfei Lin, Jiawei Han, Chen-Yu Lee, and Tomas Pfister. 2025. ReasoningBank: Scaling Agent Self-Evolving with Reasoning Memory. arXiv:2509.25140 [cs.AI]

[19] John Schulman, Filip Wolski, Prafulla Dhariwal, Alec Radford, and Oleg Klimov. 2017. Proximal Policy Optimization Algorithms. arXiv:1707.06347 [cs.LG]

[20] Zhihong Shao, Peiyi Wang, Qihao Zhu, Runxin Xu, Junxiao Song, Xiao Bi, Haowei Zhang, Mingchuan Zhang, Y. K. Li, Y. Wu, and Daya Guo. 2024. DeepSeek-Math: Pushing the Limits of Mathematical Reasoning in Open Language Models. arXiv:2402.03300 [cs.CL]

[21] Junhao Shen, Teng Zhang, Xiaoyan Zhao, and Hong Cheng. 2026. Dynamic Skill Lifecycle Management for Agentic Reinforcement Learning. arXiv:2605.10923 [cs.LG]

[22] Yaorui Shi, Yuxin Chen, Zhengxi Lu, Yuchun Miao, Shugui Liu, Qi Gu, Xunliang Cai, Xiang Wang, and An Zhang. 2026. Skill1: Unified Evolution of Skill-Augmented Agents via Reinforcement Learning. arXiv:2605.06130 [cs.AI]

[23] Noah Shinn, Federico Cassano, Ashwin Gopinath, Karthik Narasimhan, and Shunyu Yao. 2023. Reflexion: Language Agents with Verbal Reinforcement Learning. In *Advances in Neural Information Processing Systems*, Vol. 36. Curran Associates, Inc., 8634–8652. doi:10.52202/075280-0377

[24] Mohit Shridhar, Xingdi Yuan, Marc-Alexandre Côté, Yonatan Bisk, Adam Trischler, and Matthew Hausknecht. 2021. ALFWorld: Aligning Text and Embodied Environments for Interactive Learning. In *The Ninth International Conference on Learning Representations*. https://openreview.net/forum?id=0IOX0YcCdTn

[25] Shuzheng Si, Haozhe Zhao, Yu Lei, Qingyi Wang, Dingwei Chen, Zhitong Wang, Zhenhailong Wang, Kangyang Luo, Zheng Wang, Gang Chen, Fanchao Qi, Minjia Zhang, and Maosong Sun. 2026. From Context to Skills: Can Language Models Learn from Context Skillfully? arXiv:2604.27660 [cs.AI]

[26] Hui-Ze Tan, Xiao-Wen Yang, Hao Chen, Jie-Jing Shao, Yi Wen, Yuteng Shen, Weihong Luo, Xiku Du, Lan-Zhe Guo, and Yu-Feng Li. 2026. Hindsight Credit Assignment for Long-Horizon LLM Agents. arXiv:2603.08754 [cs.LG]

[27] Xiangru Tang, Tianrui Qin, Tianhao Peng, Ziyang Zhou, Daniel Shao, Tingting Du, Xinming Wei, Peng Xia, Fang Wu, He Zhu, Ge Zhang, Jiaheng Liu, Xingyao Wang, Sirui Hong, Chenglin Wu, Hao Cheng, Chi Wang, and Wangchunshu Zhou. 2025. Agent KB: Leveraging Cross-Domain Experience for Agentic Problem Solving. arXiv:2507.06229 [cs.CL]

[28] Songjun Tu, Chengdong Xu, Qichao Zhang, Yaocheng Zhang, Xiangyuan Lan, Linjing Li, Dong Li, and Dongbin Zhao. 2026. Dynamic Dual-Granularity Skill Bank for Agentic RL. arXiv:2603.28716 [cs.AI]

[29] Hanlin Wang, Chak Tou Leong, Jiashuo Wang, Jian Wang, and Wenjie Li. 2025. SPA-RL: Reinforcing LLM Agents via Stepwise Progress Attribution. arXiv:2505.20732 [cs.CL]

[30] Jiongxiao Wang, Qiaojing Yan, Yawei Wang, Yijun Tian, Soumya Smruti Mishra, Zhichao Xu, Megha Gandhi, Panpan Xu, and Lin Lee Cheong. 2025. Reinforcement Learning for Self-Improving Agent with Skill Library. arXiv:2512.17102 [cs.AI]

[31] Peiyi Wang, Lei Li, Zhihong Shao, Runxin Xu, Damai Dai, Yifei Li, Deli Chen, Yu Wu, and Zhifang Sui. 2024. Math-Shepherd: Verify and Reinforce LLMs Step-by-step without Human Annotations. In *Proceedings of the 62nd Annual Meeting of the Association for Computational Linguistics (Volume 1: Long Papers)*. Association for Computational Linguistics, 9426–9439. doi:10.18653/v1/2024.acl-long.510

[32] Prince Zizhuang Wang and Shuli Jiang. 2026. SLEA-RL: Step-Level Experience Augmented Reinforcement Learning for Multi-Turn Agentic Training. arXiv:2603.18079 [cs.LG]

[33] Quan Wei, Siliang Zeng, Chenliang Li, William Brown, Oana Frunza, Wei Deng, Anderson Schneider, Yuriy Nevmyvaka, Yang Katie Zhao, Alfredo Garcia, and Mingyi Hong. 2025. Reinforcing Multi-Turn Reasoning in LLM Agents via Turn-Level Reward Design. arXiv:2505.11821 [cs.LG]

[34] Rong Wu, Xiaoman Wang, Jianbiao Mei, Pinlong Cai, Daocheng Fu, Cheng Yang, Licheng Wen, Xuemeng Yang, Yufan Shen, Yuxin Wang, and Botian Shi. 2025. EvolveR: Self-Evolving LLM Agents through an Experience-Driven Lifecycle. arXiv:2510.16079 [cs.CL]

[35] Zhiheng Xi, Chenyang Liao, Guanyu Li, Yajie Yang, Wenxiang Chen, Zhihao Zhang, Binghai Wang, Senjie Jin, Yuhao Zhou, Jian Guan, Wei Wu, Tao Ji, Tao Gui, Qi Zhang, and Xuanjing Huang. 2025. AgentPRM: Process Reward Models for LLM Agents via Step-Wise Promise and Progress. arXiv:2511.08325 [cs.CL]

[36] Peng Xia, Jianwen Chen, Hanyang Wang, Jiaqi Liu, Kaide Zeng, Yu Wang, Siwei Han, Yiyang Zhou, Xujiang Zhao, Haifeng Chen, Zeyu Zheng, Cihang Xie, and Huaxiu Yao. 2026. SkillRL: Evolving Agents via Recursive Skill-Augmented Reinforcement Learning. arXiv:2602.08234 [cs.LG]

[37] Sikuan Yan, Xiufeng Yang, Zuchao Huang, Ercong Nie, Zifeng Ding, Zonggen Li, Xiaowen Ma, Jinhe Bi, Kristian Kersting, Jeff Z. Pan, Hinrich Schuetze, Volker Tresp, and Yunpu Ma. 2026. Memory-R1: Enhancing Large Language Model Agents to Manage and Utilize Memories via Reinforcement Learning. In *Proceedings of the 64th Annual Meeting of the Association for Computational Linguistics (Volume 1: Long Papers)*. Association for Computational Linguistics, 12805–12825. doi:10.18653/v1/2026.acl-long.583

[38] Yutao Yang, Junsong Li, Qianjun Pan, Bihao Zhan, Yuxuan Cai, Lin Du, Jie Zhou, Kai Chen, Qin Chen, Xin Li, et al. 2026. AutoSkill: Experience-Driven Lifelong Learning via Skill Self-Evolution. arXiv:2603.01145 [cs.AI]

[39] Zhilin Yang, Peng Qi, Saizheng Zhang, Yoshua Bengio, William W. Cohen, Ruslan Salakhutdinov, and Christopher D. Manning. 2018. HotpotQA: A Dataset for Diverse, Explainable Multi-hop Question Answering. In *Proceedings of the 2018 Conference on Empirical Methods in Natural Language Processing*. 2369–2380. doi:10.18653/v1/D18-1259

[40] Shunyu Yao, Howard Chen, John Yang, and Karthik Narasimhan. 2022. WebShop: Towards Scalable Real-World Web Interaction with Grounded Language Agents. In *Advances in Neural Information Processing Systems*, Vol. 35. 20744–20757. doi:10.52202/068431-1508

[41] Shunyu Yao, Jeffrey Zhao, Dian Yu, Nan Du, Izhak Shafran, Karthik Narasimhan, and Yuan Cao. 2023. ReAct: Synergizing Reasoning and Acting in Language Models. In *The Eleventh International Conference on Learning Representations*. https://openreview.net/forum?id=WE_vluYUL-X

[42] Yanwei Yue, Boci Peng, Xuanbo Fan, Jiaxin Guo, Qiankun Li, and Yan Zhang. 2026. Mem-T: Densifying Rewards for Long-Horizon Memory Agents. arXiv:2601.23014 [cs.LG]

[43] Yunpeng Zhai, Shuchang Tao, Cheng Chen, Anni Zou, Ziqian Chen, Qingxu Fu, Shinji Mai, Li Yu, Jiaji Deng, Zouying Cao, Zhaoyang Liu, Bolin Ding, and Jingren Zhou. 2025. AgentEvolver: Towards Efficient Self-Evolving Agent System. arXiv:2511.10395 [cs.LG]

[44] Guibin Zhang, Muxin Fu, and Shuicheng Yan. 2025. MemGen: Weaving Generative Latent Memory for Self-Evolving Agents. arXiv:2509.24704 [cs.CL]

[45] Hanrong Zhang, Shicheng Fan, Henry Peng Zou, Yankai Chen, Zhenting Wang, Jiayu Zhou, Chengze Li, Wei-Chieh Huang, Yifei Yao, Kening Zheng, Xue Liu, Xiaoxiao Li, and Philip S. Yu. 2026. CoEvoSkills: Self-Evolving Agent Skills via Co-Evolutionary Verification. arXiv:2604.01687 [cs.AI]

[46] Mingda Zhang, Tiesunlong Shen, Haoran Luo, Wenjin Liu, Zikai Xiao, Erik Cambria, and Xiaoying Tang. 2026. SkillFlow: Flow-Driven Recursive Skill Evolution for Agentic Orchestration. arXiv:2605.14089 [cs.AI]

[47] Shengtao Zhang, Jiaqian Wang, Ruiwen Zhou, Junwei Liao, Yuchen Feng, Zhuo Li, Yujie Zheng, Weinan Zhang, Ying Wen, Zhiyu Li, Feiyu Xiong, Yutao Qi, Bo Tang, and Muning Wen. 2026. MemRL: Self-Evolving Agents via Runtime Reinforcement Learning on Episodic Memory. arXiv:2601.03192 [cs.CL]

[48] Wenyuan Zhang, Xinghua Zhang, Haiyang Yu, Shuaiyi Nie, Bingli Wu, Juwei Yue, Tingwen Liu, and Yongbin Li. 2026. ExpSeek: Self-Triggered Experience Seeking for Web Agents. arXiv:2601.08605 [cs.CL]

[49] Xiaoying Zhang, Zichen Liu, Yipeng Zhang, Xia Hu, and Wenqi Shao. 2026. RetroAgent: From Solving to Evolving via Retrospective Dual Intrinsic Feedback. arXiv:2603.08561 [cs.AI]

[50] Zhiwei Zhang, Yudi Lin, Nikki Lijing Kuang, Linlin Wu, Xiaomin Li, Songtao Liu, and Fenglong Ma. 2026. Co-Evolving Skill Generation and Policy Optimization. arXiv:2606.08755 [cs.CL]

[51] Andrew Zhao, Daniel Huang, Quentin Xu, Matthieu Lin, Yong-Jin Liu, and Gao Huang. 2024. ExpeL: LLM Agents Are Experiential Learners. *Proceedings of the AAAI Conference on Artificial Intelligence* 38, 17 (2024), 19632–19642. doi:10.1609/aaai.v38i17.29936

[52] Han Zhou, Xingchen Wan, Ivan Vulić, and Anna Korhonen. 2025. Agentic Policy Optimization via Instruction-Policy Co-Evolution. arXiv:2512.01945 [cs.LG]

[53] Zijian Zhou, Ao Qu, Zhaoxuan Wu, Sunghwan Kim, Alok Prakash, Daniela Rus, Jinhua Zhao, Bryan Kian Hsiang Low, and Paul Pu Liang. 2025. MEM1: Learning to Synergize Memory and Reasoning for Efficient Long-Horizon Agents. arXiv:2506.15841 [cs.CL]

[54] Jiapeng Zhu, Jianxiang Yu, Yibo Zhao, Chengcheng Han, Qi Gu, Xunliang Cai, Xiang Li, and Weining Qian. 2026. Skill0.5: Joint Skill Internalization and Utilization for Out-of-Distribution Generalization in Agentic Reinforcement Learning. arXiv:2605.28424 [cs.CL]

## A Training Parameter Settings

Table 5 reports the training and rollout settings for ALFWorld and WebShop. Both environments use the same Strategy-to-Tactics disclosure interface, GRPO objective, exact-title retrieval protocol, and prompt-length budget. ALFWorld keeps a longer history for multi-step household tasks, while WebShop uses a shorter horizon and a smaller history window for product search.

**Table 5: Training and rollout settings.**

| Parameter | ALFWorld 4B | ALFWorld 8B | WebShop |
|---|---|---|---|
| Backbone | Qwen3-4B-Instruct-2507 | Qwen3-8B | Qwen2.5-7B-Instruct |
| Policy objective | GRPO objective (max.) | GRPO objective (max.) | GRPO objective (max.) |
| Training batch size | 16 | 16 | 16 |
| Max prompt length | 8192 | 8192 | 8192 |
| Max response length | 512 | 512 | 512 |
| Learning rate | $1 \times 10^{-6}$ | $1 \times 10^{-6}$ | $1 \times 10^{-6}$ |
| KL-divergence coefficient | 0.01 | 0.01 | 0.01 |
| Rollout temperature | 0.4 | 0.4 | 0.4 |
| GRPO group size | 8 | 8 | 8 |
| Environment max steps | 50 | 50 | 15 |
| History length | 50 | 50 | 2 |
| Actor mini-batch size | 64 | 128 | 64 |
| Actor micro-batch / GPU | 4 | 4 | 2 |
| Log-prob micro-batch / GPU | 4 | 4 | 2 |
| Training resources | 1 node × 8 GPUs | 2 nodes × 8 GPUs | 1 node × 8 GPUs |

The two ALFWorld runs differ mainly in actor batching and compute allocation. The 8B run uses two nodes and a larger actor mini-batch, while keeping the same rollout temperature, group size, response budget, and environment horizon as the 4B run. WebShop shares the training batch size and learning rate, but uses smaller actor micro-batches under its shorter environment horizon.

## B Skill-Bank Evolution Parameters and Rules

Table 6 reports the main settings for Strategy selection, retrieval-event credit assignment, and skill-bank evolution.

**Table 6: Strategy selection, retrieval-event credit, and skill-bank evolution settings.**

| Parameter | ALFWorld | WebShop |
|---|---|---|
| Embedding model | Qwen3-Embedding-0.6B | Qwen3-Embedding-0.6B |
| Visible Strategy entries $K$ | 3 | 3 |
| Tactic titles per Strategy $L$ | 10 | 5 |
| Strategist model | Sonnet 4.5 | Sonnet 4.5 |
| Analyst model | Sonnet 4.5 | Sonnet 4.5 |
| Analyst-score coefficient $\lambda$ | 1.0 | 1.0 |
| Retrieval warm-up $T_{\text{warmup}}$ | 20 | 40 |
| Warm-up exact-match bonus | 3.0 | 3.0 |
| Post-warm-up exact-match bonus | 0.5 | 0.5 |
| Target retrieval-count range | 1–2 | 1–2 |
| Over-retrieval penalty | 0.5 | 0.5 |
| Strategy utility EMA $\gamma_{\text{stra}}$ | 0.1 | 0.1 |
| Tactic utility EMA $\gamma_{\text{tac}}$ | 0.15 | 0.15 |
| Task-similarity weight $\alpha$ | 0.7 | 0.7 |
| UCB coefficient $\eta$ | 0.2 | 0.2 |
| Deduplication threshold | 0.85 | 0.8 |
| Probation window $T_{\text{prob}}$ | 10 | 20 |
| Effectiveness threshold $\theta_{\text{effect}}$ | 0.5 | 0.2 |
| Deprecation threshold $\theta_{\text{dep}}$ | 0.2 | 0.15 |
| Minimum evidence count $n_{\text{min}}$ | 5 | 20 |
| Strategy update threshold $\theta_{\text{refine}}$ | 0.15 | 0.15 |
| Minimum Strategy update count | 3 | 3 |

STRIDE embeds the task instruction and each Strategy's `when_to_apply` field, then ranks Strategies and child Tactics by relevance, utility, and exploration. The prompt keeps the top $K$ Strategy entries and up to $L$ child Tactic titles under each entry. Warm-up rewards exact-title retrieval; after warm-up, the exact-match bonus is reduced and Analyst feedback becomes the main retrieval signal. Skill-bank evolution uses accumulated rollout evidence to create missing Tactics, rewrite low-utility or overbroad Strategies, and prune entries only after the minimum evidence count is reached.

## C Baseline Reproduction Note

At the time of our experiments, ProactAgent [5] had no publicly available code, and its paper did not report results for the model–dataset configurations used here. We therefore omit it from the quantitative comparison.

## D Case Analysis

Figure 6 shows how a retrieved Tactic changes the next local decision in ALFWorld. Each example contrasts the retrieved path with an illustrative no-retrieval branch at the same decision point. The selected Strategy maintains the task-level plan, while the retrieved Tactic addresses the obstacle at the current state.

The three patterns differ in timing. Procedural error prevention acts before an incorrect action is taken. Prior-guided localization narrows the search while several actions remain plausible. Stuck-state recovery expands the search after likely locations have been exhausted. Together, the cases show that Tactics support both failure prevention and recovery without replacing persistent Strategy guidance.

## E Prompt Templates

The prompts implement the same Strategy-to-Tactics interface described in the main paper. Acting prompts keep the selected Strategies in context and show only Tactic titles. Post-rollout prompts use retrieval traces for Strategy and Tactic creation and update, as well as Analyst scoring. In the templates, *task-level skill* denotes a Strategy, while *step lesson* and *step skill* denote a Tactic; these implementation field names are retained verbatim.

*Acting with progressive disclosure.* The two prompts below are the environment-facing acting prompts. In both, the Tactician outputs either one admissible environment action or one exact-title retrieval request. Full Tactic content stays outside the context until the request is issued.

**ALFWorld Acting Prompt**

You are an ALFWorld embodied-task agent.
**Task:** [task description]
Step [current step] (you have taken [step count] prior steps).
**Recent history (observation → action):** [action history]
**Current observation:** [current observation]
**Admissible actions:** [admissible actions]
**Past task-level experience** (general strategy; may be empty or imperfect—use only when it aligns with the current observation): [retrieved task-level skills and child step-title menu]
**Output format:** Output EXACTLY one block, in this order, and nothing else:

```
<|think|>brief reasoning over the observation, history, and any relevant
experience above<|/think|>
<action>one admissible action, copied verbatim from the list above</
action>
```

Rules: pick exactly one admissible action and copy it character-for-character; use one `<|think|>` block and one `<action>` block per turn, with no text outside the tags.
**Optional retrieval instruction:** The task-level experience above lists, under each entry, a menu of retrievable step lessons with exact titles. You may replace `<action>` with `<retrieve>` to pull one lesson by its exact title. Its content appears in the next turn as a [Retrieved Step] block; you then produce <|think|>+<action>.
**Alternative output:**

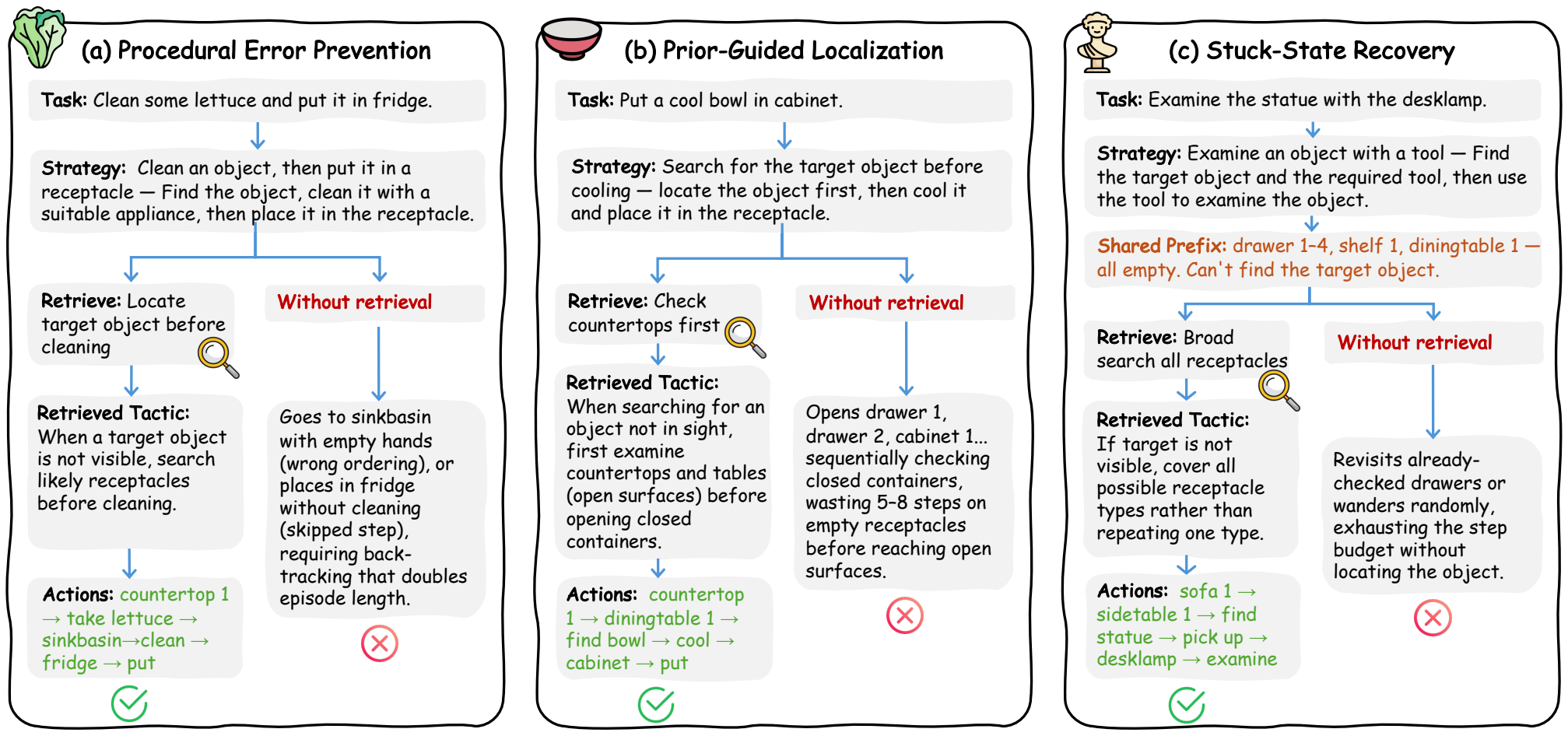

**Figure 6: Representative Tactic retrievals on ALFWorld. The three cases show how local guidance preserves action order, narrows search, or expands search after likely locations are exhausted. Illustrative no-retrieval branches show the corresponding failures.**

```
<|think|>which listed step lesson would help here and why<|/think|>
<retrieve>the exact title, copied verbatim from the menu</retrieve>
```

Use <retrieve> when a sub-goal just started, an unexpected observation broke the plan, the same action just failed or was a no-op, or one menu title clearly matches the current situation. Do not retrieve when the next action is obvious, no menu title matches, or the same lesson was already pulled in the episode. Copy the title character-for-character, and use exactly one <|think|> block plus one <retrieve> block, with no text outside the tags.

The WebShop prompt keeps the same disclosure interface but changes the observation and action fields for product search. The retrieval action is consistent across environments, while action validity is determined by each simulator.

**WebShop Acting Prompt**

You are an expert autonomous agent operating in the WebShop e-commerce environment.
**Your task is to:** [task description].
**Retrieved Relevant Experience:** [retrieved task-level skills and child step-title menu]
**Current Progress:** Prior to this step, you have already taken [step count] step(s). Below are the most recent observations and the corresponding actions you took: [action history]
You are now at step [current step] and your current observation is: [current observation].
**The admissible actions in the current situation are:** [available actions]
Now it is your turn to take one action for the current step. You should first reason step-by-step about the current situation, then think carefully about which admissible action best advances the shopping goal. This reasoning process MUST be enclosed within <|think|> <|/think|> tags. Once finished, choose an admissible action for the current step and present it within <action> </action> tags.
**Format:**

```
<|think|>your reasoning<|/think|>
<action>your chosen admissible action</action>
```

**Optional retrieval instruction.** The task-level experience above lists, under each entry, a menu of retrievable step lessons with exact titles. Instead of taking an <action> this turn, you may pull one lesson by its exact title. Its content appears in the next turn as a [Retrieved Step] block; you then reason and take an action.
**Alternative output:**

```
<|think|>which listed step lesson would help here and why<|/think|>
<retrieve>the exact title, copied verbatim from the menu</retrieve>
```

Use <retrieve> when a sub-goal just started, the last action failed or was a no-op, an unexpected observation broke the plan, or one menu title clearly matches the current situation. Do not retrieve when the next action is obvious, no title matches, or the same lesson was already pulled. Copy the title character-for-character, and use exactly one <|think|> block plus one <retrieve> block, with no text outside the tags and no <action> tag.

*Updating the skill bank.* After each training batch, the Strategist uses retrieval traces and task outcomes for Strategy and Tactic creation and update. The generation prompt below compares a failed trajectory with a successful one from the same task pattern, identifies their first consequential divergence, and writes the missing local procedure as a Tactic. The Strategist creates a new Strategy for a distinct task pattern.

**Skill Generation Prompt**

You are given one FAILED trajectory and one SUCCESSFUL trajectory from the same task. Compare them to identify where the failed one first diverged from the correct approach, then extract a reusable step-level lesson.
Trajectory format note: a single environment step that triggered skill retrieval is shown as two consecutive turns: Turn $N$ (retrieve), containing the agent's <retrieve> query, followed by Turn $N+1$ (act), whose observation begins with [Retrieved step-level skills] and lists the returned skills.
When writing a new STEP_REFLECTION, prefer producing a fix that the retrieved skills did not cover.
**Task:** [task instruction]
**Task Type:** [task type]
**SUCCESSFUL trajectory** (reference—shows the correct approach): [successful trajectory]
**FAILED trajectory** (analyze this): [failed trajectory]
**Candidate TASK-LEVEL skills:**
 [skill id]
 title: [task skill title]
 when_to_apply: [task skill trigger]
 principle: [task skill principle]
The STEP_REFLECTION must not repeat or paraphrase these task skills. It should be a specific local fix for the exact step where the failed trajectory deviated from the correct approach.
Output the following in order using the exact section headers:

(1) FIRST_ERROR_STEP: $N$, where $N$ is the 1-based turn number in the failed trajectory where the agent first diverged. Use 0 if unclear.
(2) STEP_REFLECTION: one JSON object with exactly **title**, **principle**, and **when_to_apply**. The skill must describe one local fix, not a multi-step procedure.
For ALFWorld, the title is at most eight words and names the reusable lesson; the principle uses class words rather than object IDs; when_to_apply is one short situation-type sentence.
For WebShop, the title is 3–8 words naming only the observable local state or failure symptom; it must not reveal the remedy through verbs such as *search*, *verify*, *resolve*, *select*, or *correction*. Its trigger describes the local shopping state rather than the original product request, and its principle gives the executable local correction without product IDs, option values, prices, or meta-retrieval instructions. Do not clone an existing retrieved skill.
(3) PARENT_TASK_SKILL_IDS: a JSON array containing exactly one candidate skill_id, or an empty array. Never output multiple ids. For WebShop, pick exactly one candidate whose when_to_apply covers the purchase-need pattern and whose principle can own the step lesson; sharing with other tasks can happen later only through semantic step deduplication. Output [] when every candidate is clearly the wrong scope; the trainer will then mint a fresh task skill from the matching successful trajectory and attach the step to it.

**Output format:**

```
FIRST_ERROR_STEP: N
STEP_REFLECTION: <single JSON object>
PARENT_TASK_SKILL_IDS: <JSON array of skill_id strings>
```

Updates use evidence accumulated across rollouts. The prompt below updates Strategy guidance when its scope is inaccurate or too broad and detaches a coherent subset of Tactics when they follow a different task pattern.

**Task-Skill Refinement Prompt**

You are rewriting a TASK-level skill whose retrieval utility is low. The skill is retrieved frequently but leads to poor outcomes, which suggests its description—title, principle, or when_to_apply—is misleading or too vague. Your job is to revise it so the description accurately captures the task pattern, applicability boundary, and high-level strategy that the child step skills below collectively fall under. A TASK skill is not a page-state fix; concrete local tactics stay in the child step skills.
**SEMANTIC ANCHOR:** the revised skill must remain about the same task category as the original. Only correct inaccuracies or sharpen vague wording; do not drift into a different task pattern.
**Current task skill:**
title: [title]
principle: [principle]
when_to_apply: [trigger]
fixed category, if applicable: [webshop category]
**Child step skills under this task:**
[step id]: title=[title]; principle=[principle];
when_to_apply=[trigger]
Rewrite the task skill. Output a single JSON object with exactly **title**, **principle**, and **when_to_apply**; for WebShop also include **detach_step_ids**, drawn only from the listed children, or `[]` if all remain compatible.
**Output format:**

```
TASK_SKILL: <single JSON object>
```

*Scoring retrieval events.* The Analyst prompt below scores each retrieval event for necessity and relevance. Its timing check determines whether the state required local guidance, while its title-validity check determines whether the requested title appeared in the visible Tactic-title menu and matched the local situation. The resulting GOOD/BAD label is used during training to shape the corresponding retrieval decision; it is not an independent human validation label. Task outcomes separately update skill utilities.

**Hindsight PRM Prompt**

**Task:** [task instruction]
**Available step skill titles:**
- [title 1]
- [title 2]
- [title 3]

**Trajectory:**
[Turn 1]
Observation: [observation]
Action: [action]
· · ·
[Retrieve Step #$N$]
Observation before retrieval: [observation]
Selected title: [retrieved title]
Action after retrieval: [following action]
**Final outcome:** [success/failure or score]
For each [Retrieve Step #$N$], judge two things.
First, **timing**: was the agent genuinely stuck or uncertain about what to do next? GOOD timing means the next action is non-obvious from the observation alone, or the agent has attempted multiple actions without progress. BAD timing means the correct next action is already clear and untried.
Second, **title validity**: is the selected title in the available skill list and relevant to the current situation?
**GOOD** (+1) = timing appropriate and title valid/relevant. **BAD** ($-1$) = bad timing, invalid title, or irrelevant title.
**Return JSON only:**

```
{
  "retrieval_evaluations": [
    {"step_number": <int>, "judgment": "GOOD"/"BAD"}
  ]
}
```